\PassOptionsToPackage{numbers}{natbib}
\documentclass{article}

\usepackage[preprint]{neurips_2024}

\usepackage[utf8]{inputenc} 
\usepackage[T1]{fontenc}    
\usepackage{hyperref}       
\usepackage{url}            
\usepackage{booktabs}       
\usepackage{amsfonts}       
\usepackage{nicefrac}       
\usepackage{microtype}      
\usepackage{xcolor}         

\usepackage{amsthm}
\newtheorem{theorem}{Theorem}[section]
\newtheorem{corollary}{Corollary}[theorem]

\usepackage{amsmath}
\usepackage{bm}
\usepackage{gensymb}
\usepackage{float}
\usepackage{graphicx}
\usepackage{subfig}

\usepackage{caption}
\usepackage{subcaption}
\usepackage{multirow}
\usepackage{mathtools}

\title{CLERF: Contrastive LEaRning for Full Range Head Pose Estimation}

\author{ Ting-Ruen Wei \\
  Santa Clara University \\
  Santa Clara, CA \\
   \And
  Haowei Liu \\
  Santa Clara University \\
  Santa Clara, CA \\
   \And
  Huei-Chung Hu \\
  DOCOMO Innovations \\
  Sunnyvale, CA \\
  \And
  Xuyang Wu \\
  Santa Clara University \\
  Santa Clara, CA\\
  \And 
  Yi Fang \\
  Santa Clara University \\
  Santa Clara, CA \\
  \And
  Hsin-Tai Wu \\
  DOCOMO Innovations \\
  Sunnyvale, CA \\
  }

\begin{document}

\maketitle

\begin{abstract}
We introduce a novel framework for representation learning in head pose estimation (HPE). Previously such a scheme was difficult due to head pose data sparsity, making triplet sampling infeasible. Recent progress in 3D generative adversarial networks (3D-aware GAN) has opened the door for easily sampling triplets (anchor, positive, negative). We perform contrastive learning on extensively augmented data including geometric transformations and demonstrate that contrastive learning allows networks to learn genuine features that contribute to accurate HPE. On the other hand, we observe that existing HPE works struggle to predict head poses as accurately when test image rotation matrices are slightly out of the training dataset distribution. Experiments show that our methodology performs on par with state-of-the-art models on standard test datasets and outperforms them when images are slightly rotated/ flipped or full range head pose. To the best of our knowledge, we are the first to deliver a true full range HPE model capable of accurately predicting any head pose including upside-down pose. Furthermore, we compared with other existing full-yaw range models and demonstrated superior results.
\end{abstract}

\section{Introduction}
In the expansive landscape of machine learning, contrastive learning has marked its territory as a pivotal technique, particularly within the unsupervised learning paradigm \cite{chen2020simple}. It operates on a simple yet effective principle: teaching models to recognize similar and dissimilar instances, and leveraging large datasets to enhance model understanding and performance. While its application has been widely applied in numerous aspects of computer vision, the venture into Full Range (FR) Head Pose Estimation (HPE) using contrastive learning would be entering uncharted territory. Please note currently available FR models are not in our standard full range because none of them are capable of handling, for example, upside-down head poses. HPE is a complex and crucial task in computer vision \cite{hempel2024toward, cao2020vectorbased, zhou2020whenet, cobo2024representation}, which aims to accurately determine the orientation of a person's head, and the task is important for understanding human behavior and intentions in various applications, from augmented and virtual reality environments to safety systems in vehicles and interactive robotics. 

The challenge in applying contrastive learning to FRHPE lies in the sparsity of head poses in the 3D space. It is extremely rare to find an anchor-positive (another head facing the same direction) in the full range 3D space. If we allow anchor-positives to be twenty degrees apart from the anchor at the maximum, the probability of finding an anchor-positive is under 0.0002 ($\frac{1}{18^3}$), exemplifying the difficulty of applying contrastive learning.

While many existing works in HPE focus on the limited frontal range \cite{yang2019fsa, huang2020improving, Ruiz_2018_CVPR_Workshops, dhingra2022lwposr, hsu2018quatnet, cao2020vectorbased, dai2020rankpose, zhang2023tokenhpe}, with yaw in the range of -90 to 90 degrees, research in FR models \cite{zhou2020whenet, hempel2024toward} are underdeveloped. While it remains a challenge to curate a dataset that covers the FR, which we define as the range of -180 to 180 degrees for all yaw, pitch, and roll (refer to \cite{enwiki:1223812775}), the FR capability extends coverage from the limited frontal range and is particularly significant in HPE of sports and acrobatic actions. Additionally, we observe that many existing models are sensitive to slight transformations of the test images: minor rotation and/or flip.

To tackle these challenges, we propose a framework to train with contrastive learning a FR model (CLERF) that demonstrates competitive performance on not only original test images but also on slightly augmented versions and other angles that existing models struggle on. Specifically, CLERF generates a synthetic head image with the same yaw and pitch as a real image and geometrically transforms the generated head image to have the exact head orientation as the real image. The process guarantees a positive pair thereby enabling the use of contrastive learning. The advantage of synthetic data lies in its flexibility to represent any head orientation and address a wider range including the angles that are rarely observed in real data. With geometric transformation, we can further expand the coverage to more angles. Through contrastive learning and image augmentation, CLERF aims to learn a good representation, separating neighboring angles from further ones in the FR and making more accurate head pose predictions.

Our main contribution can be summarized as follows:

\begin{itemize}
  \item We identify the advantage of 3D-aware GAN to generate anchor-positives and facilitate contrastive learning in full range head pose estimation.
  \item We perform calculations to find parameters of general geometric transformations. Such transformations allow 3D-aware GAN to synthesize positive images to match anchors and expand the head pose coverage to full range.
  \item We observe that existing models are sensitive to slight transformations of the test images.
  \item We demonstrate on par performance with state-of-the-art models on original images of the standard test sets and outperformed them on minor variants.
  \item Our model is capable of handling true FR head pose and outperforms existing full-yaw range models at the full range capability. Code will be released.
\end{itemize}

\section{Related Work}

\textbf{Head pose estimation.} Classical approaches, such as template matching and detector arrays, related to head pose estimation can be found in the survey paper \cite{hpe_article}. Deformable models \cite{1227983, DBLP:conf/cvpr/ZhuLLSL16, guo2020towards, zhu2017face} were used to create the commonly used synthetic pose dataset such as 300W-LP \cite{DBLP:conf/cvpr/ZhuLLSL16}. Head pose estimation can be divided into two categories, with and without facial landmarks. Traditionally, HPE was made when facial features are visible, with Dlib \cite{dlib09} being one of the pioneers in using face landmarks for prediction. However, such method becomes error-prone when facial landmarks are not detected, especially at large yaw angles. 
Therefore, with the advent of deep learning, researchers utilized Convolutional Neural Networks (CNN) to predict the three Euler angles directly, as in HopeNet \cite{Ruiz_2018_CVPR_Workshops} and WHENet \cite{zhou2020whenet}. Nonetheless, directly applying regression on the Euler angles leads to discontinuity easily because each distinct angle can be represented by different numbers, e.g., $0\degree = 360\degree$. To avoid such discontinuity, 6DRepNet \cite{hempel2024toward}, 6DRepNet360 \cite{hempel2024toward}, and TriNet \cite{cao2020vectorbased} predict the $3 \times 2$ and $3 \times 3$ rotation matrices respectively, while still evaluating the Euler angles for comparison with other HPE models. Other works include Kuhnke et al. \cite{kuhnke2023domain}, who bridged the gap between synthetic and realistic images in head pose estimation using relative pose consistency, and Opal \cite{cobo2024representation}, which aligned the different reference systems between the training and testing datasets and proposed a generalized geodesic distance metric as the loss function. SemiUHPE \cite{zhou2024semi} applied weak-strong augmentations in a semi-supervised fashion, leveraging a large amount of unlabeled head poses. Instead of the common CNN approach, TokenHPE \cite{zhang2023tokenhpe} utilized a transformer for HPE by predicting orientation tokens.

\textbf{Head pose dataset creation.} The 300W across Large Poses (300W-LP) dataset \cite{DBLP:conf/cvpr/ZhuLLSL16} was based on 300W \cite{DBLP:conf/iccvw/SagonasTZP13}, which standardized multiple alignment datasets with 68 landmarks, including AFW \cite{DBLP:conf/cvpr/ZhuR12}, LFPW \cite{DBLP:conf/cvpr/BelhumeurJKK11}, HELEN \cite{DBLP:conf/iccvw/ZhouFCJY13}, IBUG \cite{DBLP:conf/iccvw/SagonasTZP13}, and XM2VTS \cite{Messer1999XM2VTSDBTE}. The Euler angles labeled are extracted from the rotation matrix estimated from its 3D Dense Face Alignment, which applies the morphable model 3DMM \cite{1227983} to obtain the standard and rotated 3D faces tailored for an image. While 300W-LP offered pose labels limited to frontal views, the CMU Panoptic Dataset \cite{Joo_2017_TPAMI} provided extensive 3D facial landmarks for multiple individuals captured from cameras spanning an entire hemisphere, which can be converted to a head pose. WHENet \cite{zhou2020whenet} pioneered techniques to transform the CMU Panoptic Dataset into a comprehensive head pose estimation dataset covering a larger range of angles.

\textbf{Contrastive Learning.} To the best of our knowledge, no prior work has utilized contrastive learning in head pose estimation, so we discuss a few contrastive learning applications in related computer vision areas. For gaze estimation, GazeCLR \cite{jindal2023contrastive} revised the NT-Xent loss \cite{chen2020simple} on multi-view images to learn gaze representation and improve the cross-domain gaze estimation performance. CRGA \cite{wang2022contrastive} utilized contrastive regression in gaze estimation for a domain adaptation task. As for hand pose estimation, PeCLR \cite{Spurr_2021_ICCV} modified the SimCLR \cite{chen2020simple} framework for equivariance contrastive learning on 3D hands and improved the performance of the original models. In face recognition, PCL \cite{liu2023pose} learned face representations by disentangling the pose from the original face in computing the contrastive loss. For human activity representation, P-HLVC \cite{schneider2022pose} leveraged human pose dynamics to learn human activities that are resistant to domain shifts. In human pose estimation, Honari et al. \cite{honari2022temporal} separated time-variant from time-invariant features and only applied contrastive learning on time-invariant features. ICON \cite{xu2023inter} enforced the consistency between individual keypoints belonging to the same category across images and also the consistency between pair relations across images in a multi-person scenario.

\begin{figure}[t]
\centering
    \includegraphics[width=\textwidth]{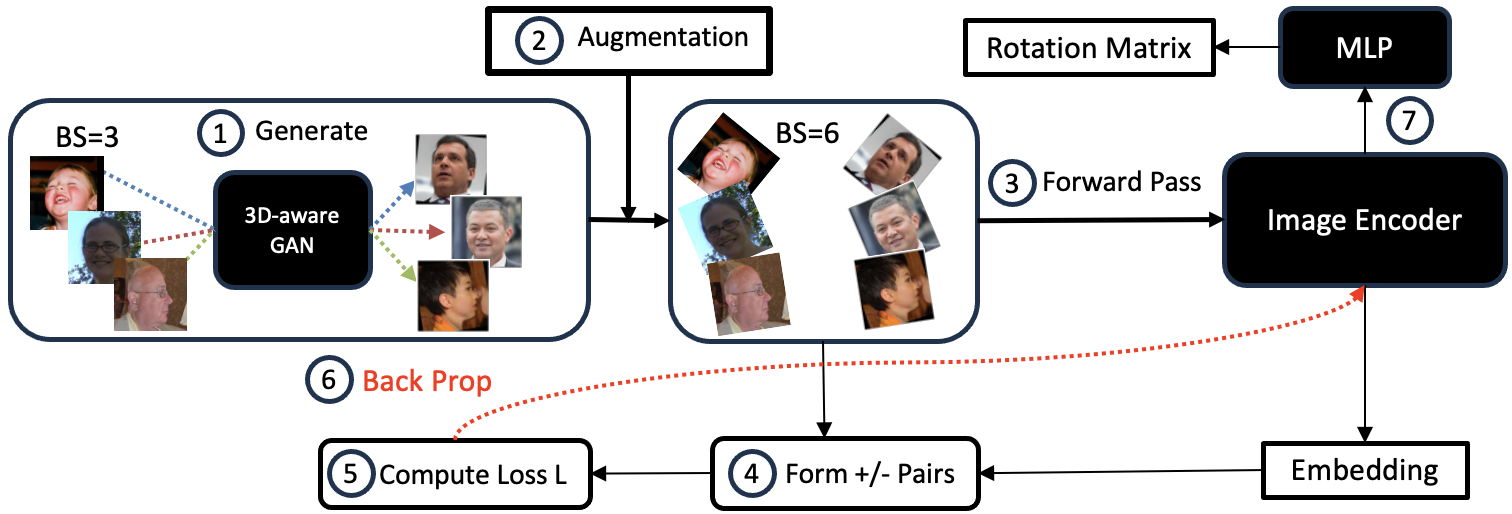}
\caption{Proposed method for contrastive learning in full range head pose representation, with a batch size of 3 instances for illustration purposes. Steps 1$\sim$6 comprise a training iteration for the representation model which is frozen in the downstream MLP training in step 7.}
\label{fig:flow}
\end{figure}

\section{Methodology}
Our proposed methodology involves three key components: anchor-positive generation to empower contrastive learning, geometric transformation to cover the full range space, and contrastive learning to strengthen the fine-grained head pose predictions. An overview is shown in Figure \ref{fig:flow}. 

\subsection{Anchor-Positive Synthetic Image Generation}
\label{subsec:synthetic}

Given a training sample, we generate with 3D-aware GAN a synthetic image that has the same yaw and pitch values and apply proper rotation to match the roll value (Figure \ref{fig:flow} step 1). With the mathematics listed in Hu et al. \cite{hu2024mathematical}, we can solve for a triad of $(yaw, pitch, \phi)$ from the rotation matrix. The rotation associated with the roll angle $\phi$ based on the rotation matrix $R \in SO(3)$ (refer to \cite{enwiki:1099134153}) representing rotation with yaw and pitch is the following:

\begin{equation}
  \label{formula:rotation}
  \begin{split}
    R_{rotate}(\phi) &= R_{extrinsic}(\phi) \times R  \\ 
    &=  \begin{bmatrix}
        cos(\phi) & sin(\phi) & 0\\
        -sin(\phi) & cos(\phi) & 0\\
        0 & 0 & 1
       \end{bmatrix} \times R 
  \end{split}
\end{equation}

where $R_{rotate}$ is the rotation matrix of the rotated image. We provide detailed formulation below.

a novel two-stage self-adaptive image alignment for robust training, a tri-grid neural volume that effectively manages the representation challenges of both the face and the back of the head, and the integration of prior knowledge from 2D image segmentation to improve 3D model training. Panohead allows the generation of a head pose facing a specific yaw and pitch, which is a crucial component in facilitating contrastive learning. As a starting dataset, we generated 25,984 images with 812 unique-looking people and 32 images each. Each set of 32 images is uniformly distributed within the yaw and pitch space of [-3.14 radians, 3 radians] and [-1.5 radians, 0.1 radians]. The rotation matrix $R$ follows the reference system of 300W-LP as below:

\begin{figure}[t]
\begin{minipage}[t]{0.32\textwidth}
    \centering
    \includegraphics[width=0.8\textwidth]{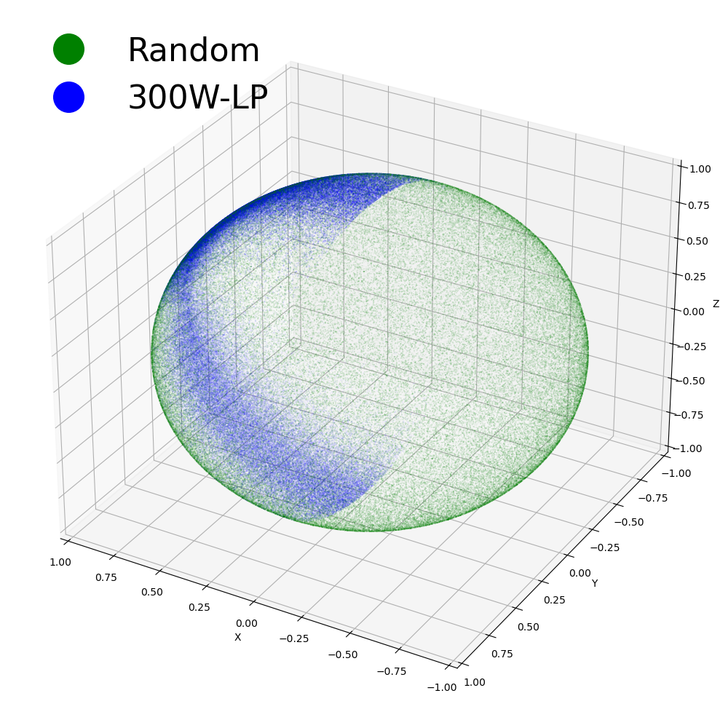}
\end{minipage}
\hfill
\begin{minipage}[t]{0.32\textwidth}
    \centering
    \includegraphics[width=0.8\textwidth]{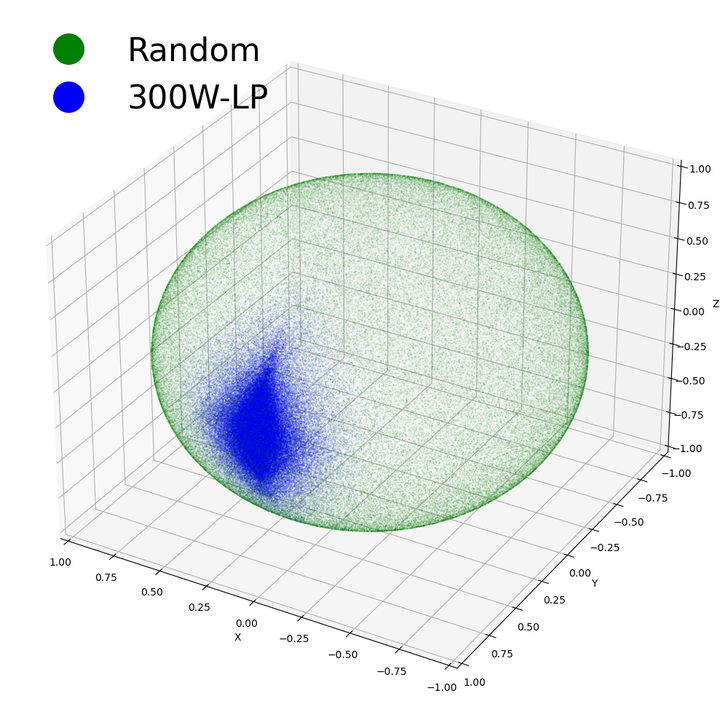}
\end{minipage}
\hfill
\begin{minipage}[t]{0.32\textwidth}
    \centering
    \includegraphics[width=0.8\textwidth]{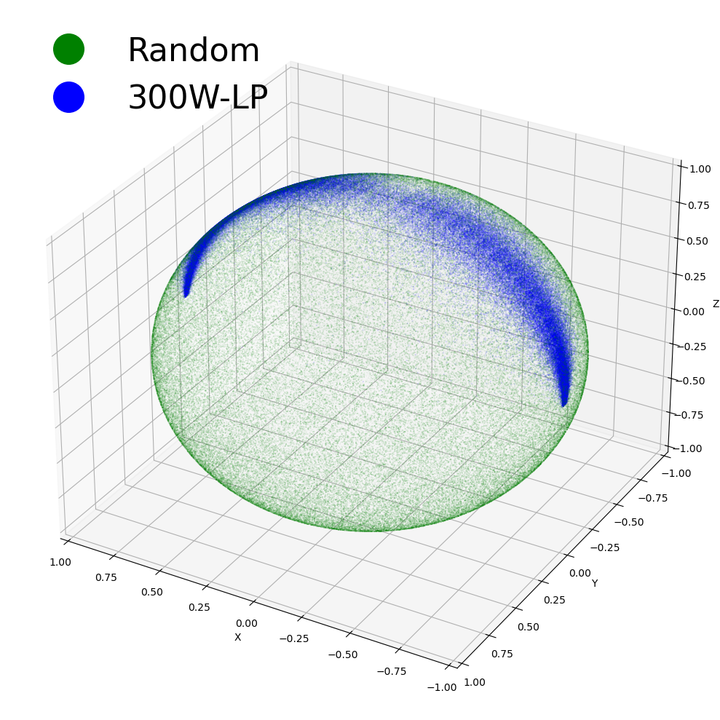}
\end{minipage}
\hfill
\begin{minipage}[t]{0.32\textwidth}
    \centering
    \includegraphics[width=0.8\textwidth]{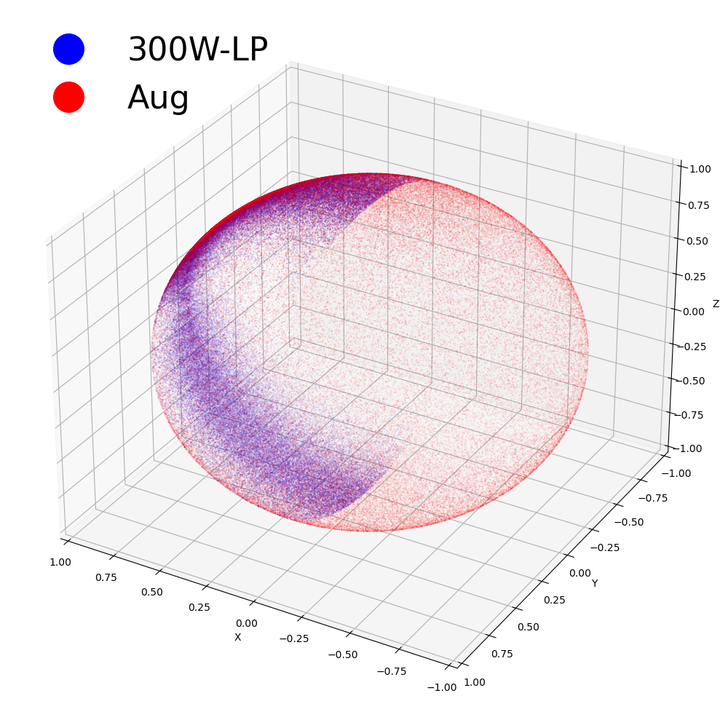}
    \mbox{(a) x axis (1, 0, 0)}
\end{minipage}
\hfill
\begin{minipage}[t]{0.32\textwidth}
    \centering
    \includegraphics[width=0.8\textwidth]{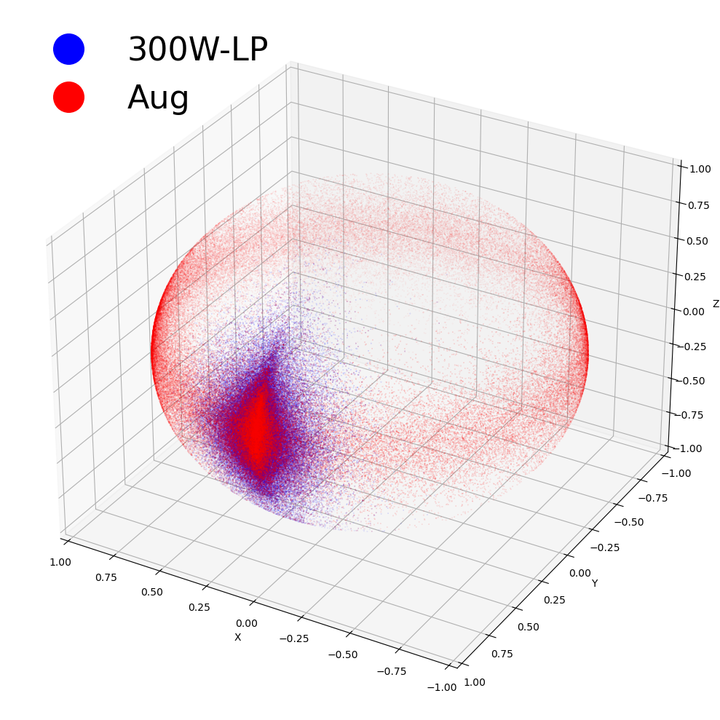}
    \mbox{(b) y axis (0, 1, 0)}
\end{minipage}
\hfill
\begin{minipage}[t]{0.32\textwidth}
    \centering
    \includegraphics[width=0.8\textwidth]{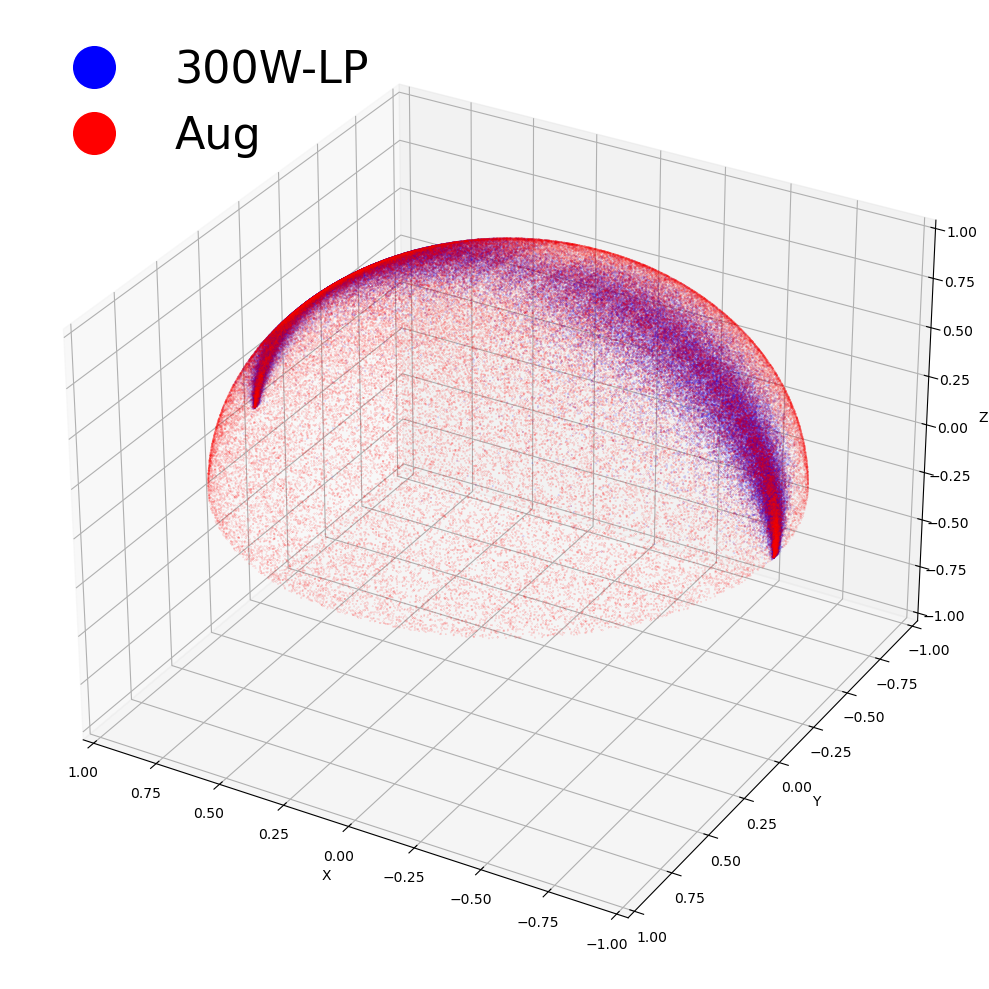}
    \mbox{(c) z axis (0, 0, 1)}
\end{minipage}
\caption{Visualization of 300W-LP dataset \cite{zhu2017face} after randomized rotation and flipping augmentations. A point on the sphere $\in \mathbb{R}^3$ is formed by projecting a rotation matrix to the 3D sphere, which multiplies unit vectors in (a) x, (b) y, and (c) z coordinate axes. The top row shows random rotation matrices (in green) along with those in the 300W-LP dataset (in blue), and the bottom row shows the augmented 300W-LP dataset (in red). The geometric transformations expand the original dataset for wider coverage.}
\label{spheres}
\end{figure}

\subsection{Geometric Transformation for Full Range HPE}
\label{subsec:headpose_aug}
The second key lies in the geometric transformation on head pose images and their corresponding rotation matrices to enable FRHPE (Figure \ref{fig:flow} step 2). Besides rotation defined in Equation \ref{formula:rotation}, we follow the mathematics in Hu et al. \cite{hu2024mathematical} and apply flipping across $L_{\theta}$ on the XY-plane with a given rotation $R \in SO(3)$ to obtain the following rotation matrix:

\begin{equation}
    \label{formula:flip}
    \begin{split}
        R_{flip}(\theta) &= Flip{\_}\theta \times R \times Flip{\_}X  \\
        &= \begin{bmatrix}
            cos(2\theta) & sin(2\theta) & 0\\
            sin(2\theta) & -cos(2\theta) & 0\\
            0 & 0 & 1
           \end{bmatrix}  \times R \times \begin{bmatrix}
            -1 & 0 & 0\\
            0 & 1 & 0\\
            0 & 0 & 1
           \end{bmatrix}    
    \end{split}
\end{equation}

where $\theta$ is the counter-clockwise angle between the positive x-axis and the line $L$ that we flip across. Figure \ref{spheres} shows the resulting distribution of the 300W-LP dataset after our rotation and flipping (in red), compared to that of random rotation matrices (in green) and the original distribution (in blue). As the 300W-LP dataset lacks head poses facing away from the camera (large yaw angles), the resulting distribution does not fully match the uniform distribution (a semi-sphere versus a sphere in Figure \ref{spheres}(c)). Nonetheless, our rotation and flipping augmentation fills the gap and transforms head poses to cover a much wider range than the original images.

\begin{theorem}
Let \(H\) be a image geometric transformation function and $A, B$ present two valid rotation matrices, then $d(A, B) = d(H(A), H(B))$.
\label{theorem1}
\end{theorem}

To maintain the guaranteed existence of a positive pair, we apply the same rotation and flipping augmentation to both images in a positive pair. Since this is a crucial condition for contrastive learning, we prove Theorem \ref{theorem1} in Appendix \ref{appendix:appendix_theoren_keep_geodesic_dist} that geometric transformations preserve the geodesic distance $d$ for any two rotation matrices $A, B \in SO(3)$, with $d$ defined as the following \cite{hempel2024toward}:

\begin{equation}
    \label{gedesic_dist}
    d(A, B) = cos^{-1}(\frac{tr(A \times B^{T})-1}{2}) 
\end{equation}

where $tr$ and $T$ represent the trace and transpose, respectively. In other words, for any chosen anchor, positive, negative triplet $(a, p, n)$, applying the same rotation and flipping augmentations results in another valid triplet $(a', p', n')$.
We present our strategy for sampling triplets for contrastive learning.

\begin{corollary}
    Strategy for sampling (anchor, positive, negative) triplets under the mathematical foundation in \cite{hu2024mathematical}:
\begin{enumerate}
\item Take an image $I_A$ as anchor and generate a synthetic image $I_P$ with rotation matrix $R' = (R_{pitch} \times R_{yaw})$.
\item For anchor images provided with rotation matrices $R$ as labels, we consider solving the corresponding triad of $(yaw, pitch, roll)$ for rotation matrix 
$R = R_{roll} \times (R_{pitch} \times R_{yaw})$. Then we rotate the image $I_P$ with $R_{roll}$ to create a positive image with rotation matrix $R_{roll} \times R' = R$. Refer to Figure \ref{fig:panohead} in Appendix \ref{sec:panohead} for examples.

\item For any existing anchor, positive, negative triplet $(a, p, n)$, applying random rotation or flipping augmentation $H$ returns a new valid triplet $(H(a), H(p), H(n))$.

\end{enumerate}

Armed with the result above, we can easily sample lots of valid triplets and their augmented versions for contrastive learning.

\end{corollary}

\subsection{Contrastive Learning}
After data augmentation, we pass the batch of images to the image encoder $E$ to obtain their embedding vectors (Figure \ref{fig:flow} step 3) and form triplets. Though the image generation step guarantees an anchor-positive, we include neighboring head orientations that have a geodesic similarity $cos(d(R, R'))$ higher than the threshold $T_{GD}$ as additional anchor-positives, where $R$ and $R'$ are the rotation matrices of the anchor and another image in the mini-batch (Figure \ref{fig:flow} step 4). Furthermore, we filter and retain the hard and semi-hard negatives that violate a margin $v$ according to the Euclidean distance between the embedding vectors. Subsequently for back-propagation (Figure \ref{fig:flow} step 6), we compute the Circle Loss \cite{sun2020circle} $L$ (Figure \ref{fig:flow} step 5) that aims to increase the similarity between positive pairs and decrease that between negative pairs, through a re-weighting process to focus on the less-optimized samples, with $L$ defined as the following:

\begin{equation}
    log(1+\sum_{i=1}^{N_p}exp(\gamma*max(0,1+m-s_i)(s_i-m)) \sum_{j=1}^{N_n}exp(-\gamma*max(0,m+s_j)(s_j-1+m)))
\end{equation}

where $\gamma$ is a scaling factor, $m$ is a relaxation margin, $s$ is the cosine similarity of the embedding vectors, $N_p$ and $N_n$ are the number of positive and negative pairs respectively.

After the representation model is trained, we freeze it and train a downstream multi-layer perceptron (MLP) to project the representation to the fine-grained head pose angles, represented by a rotation matrix (Figure \ref{fig:flow} step 7). To ensure a unitary matrix, we transform the six-dimensional MLP output to the nine-dimensional rotation matrix through the Gram-Schmidt process as implemented in 6DRepNet \cite{hempel2024toward}. With a trained representation model and its downstream MLP, we evaluate test datasets and compare against baseline models.

\section{Experiments}

\subsection{Datasets}
\textbf{Training.} 300W-LP \cite{zhu2017face} contains 122,450 images from multiple databases with faces mainly in the frontal range. We utilize PanoHead \cite{an2023panohead} as the 3D-aware GAN to generate anchor-positives for the representation model training only.

\textbf{Testing.} Following previous works, we test the models on AFLW2000 \cite{zhu2017face} and BIWI \cite{fanelli2013random}. These datasets mainly contain frontal faces with yaw in the range of [-90, 90] degrees. Therefore, to facilitate a comprehensive analysis of our full range capability, we additionally evaluated on four variants of the previous test sets: slightly-augmented (SA) AFLW2000, SA BIWI, fully-augmented (FA) AFLW2000, and FA BIWI. The SA version is obtained by rotating each image clockwise by 10 degrees and flipping it along the line that is 85 degrees counter-clockwise from the positive x-axis (an example is shown in the second row of Figure \ref{fig:cases}). As for the FA version, we randomly rotate between -180 to 180 degrees and flip along the line that is between 0 to 90 degrees counter-clockwise from the positive x-axis (an example is shown in the third row of Figure \ref{fig:cases}).

\subsection{Experimental Setup}

\begin{table}
\centering
\caption{Model performance against baseline models across six datasets. CLERF achieved on par performance with the baseline models on the original AFLW2000 and BIWI, and outperformed all baseline models when test images are slightly rotated and flipped (SA AFLW2000 and SA BIWI). In the full range, CLERF outperformed other full range models on heavily-rotated images by a large margin. FR indicates whether the model is full range, and the optimal results are highlighted in bold.}
\begin{tabular}{c|c|c|c|c|c|c|c|c|c}
 \toprule
 \multirow{2}{*}{Model} & \multirow{2}{*}{FR} & Yaw & Pitch & Roll & Mean & Yaw & Pitch & Roll & Mean\\ 
    \cmidrule(lr){3-6} \cmidrule(lr){7-10}
   & & \multicolumn{4}{|c|}{AFLW2000} & \multicolumn{4}{|c}{BIWI} \\
   \midrule
 FSA-Net \cite{yang2019fsa} & $\times$ & 4.50 & 6.08 & 4.64 & 5.07 & 4.27 & 4.96 & 2.76 & 4.00 \\
 HopeNet \cite{Ruiz_2018_CVPR_Workshops} & $\times$ & 6.47 & 6.56 & 5.44 & 6.16 & 5.17 & 6.98 & 3.39 & 5.18 \\
 TokenHPE \cite{zhang2023tokenhpe} & $\times$ & 5.44 & \textbf{4.36} & 4.08 & 4.66 & 4.51 & \textbf{3.95} & \textbf{2.71} & \textbf{3.72} \\
 6DRepNet \cite{hempel2024toward} & $\times$ & \textbf{3.27} & 4.58 & \textbf{2.98} & \textbf{3.61} & \textbf{3.23} & 5.32 & 2.78 & 3.78 \\ 
 WHENet \cite{zhou2020whenet} & $\checkmark$ & 5.11 & 6.24 & 4.92 & 5.42 & 3.99 & 4.39 & 3.06 & 3.81 \\
 6DRepNet360 \cite{hempel2024toward} & $\checkmark$ & 3.58 & 5.28 & 3.46 & 4.11 & 3.28 & 6.06 & 3.08 & 4.14  \\ 
 CLERF & $\checkmark$ & 4.22 & 6.18 & 4.67 & 5.02 & 3.57 & 4.49 & 3.13 & 3.73\\ 
 
 \midrule
   &  & \multicolumn{4}{|c|}{SA AFLW2000} & \multicolumn{4}{|c}{SA BIWI}\\ 
   \midrule
 FSA-Net \cite{yang2019fsa} & $\times$ & 18.59 & 16.02 & 17.04 & 17.22 & 7.09 & 9.42 & 6.20 & 7.57 \\
 HopeNet \cite{Ruiz_2018_CVPR_Workshops} & $\times$ & 6.44 & 9.31 & 6.08 & 7.28 & 10.80 & 10.07 & 9.32 & 10.07 \\
 TokenHPE \cite{zhang2023tokenhpe} & $\times$ & 6.29 & 7.29 & 6.56 & 6.70 & 6.84 & 7.12 & 5.19 & 6.39 \\
 6DRepNet \cite{hempel2024toward} & $\times$ & 8.41 & 8.80 & 7.68 & 8.30 & 6.45 & 7.09 & 6.79 & 6.78 \\
 WHENet \cite{zhou2020whenet} & $\checkmark$ & 13.11 & 12.88 & 15.06 & 13.68 &  9.51 & 10.99 & 9.52 & 10.00 \\
 6DRepNet360 \cite{hempel2024toward} & $\checkmark$ & 5.69 & 6.76 & 5.34 & 5.93 & 10.21 & 7.88 & 6.51 & 8.20\\ 
 CLERF & $\checkmark$ & \textbf{4.56} & \textbf{6.10} & \textbf{4.86} & \textbf{5.36} & \textbf{6.57} & \textbf{4.52} & \textbf{4.21} & \textbf{5.10} \\ 

 \midrule
   &  & \multicolumn{4}{|c|}{FA AFLW2000} & \multicolumn{4}{|c}{FA BIWI}\\ 
   \midrule
 WHENet \cite{zhou2020whenet} & $\checkmark$ & 22.04 & 23.03 & 39.82 & 28.30 & 30.77 & 22.43 & 41.65 & 31.95 \\
 6DRepNet360 \cite{hempel2024toward} & $\checkmark$ & 14.00 & 16.93 & 21.96 & 17.63 & 25.97 & 17.90 & 34.04 & 25.97\\ 
 CLERF & $\checkmark$ & \textbf{4.84} & \textbf{5.79} & \textbf{4.31} & \textbf{4.98} & \textbf{7.68} & \textbf{7.89} & \textbf{6.93} & \textbf{7.50} \\ 
 
 \bottomrule
\end{tabular}
\label{datasets}
\end{table}

\textbf{Representation Model.} We use the improved version of Swin Transformer Base model \cite{liu2022swin} with weights pre-trained on ImageNet in PyTorch and an output embedding of size 1024 as the image encoder $E$. We train it for 30 epochs on 20,000 images from 300W-LP dataset and 20,000 PanoHead-generated images with the Adam optimizer and a learning rate of $10^{-5}$ on a single NVIDIA Tesla V100 32GB GPU. We utilize the PyTorch Metric Learning library \cite{Musgrave2020PyTorchML} for triplet sampling and loss computation. For hyperparamter values, we apply $T_{GD}=0.8$ and $v=0.1$ for triplet filtering and $m=0.4$ and $\gamma=80$ for loss $L$.

\textbf{MLP.} Our downstream MLP consists of four fully connected layers of 256 units and a skip connection from the input layer to the last fully connected layer. We train the MLP for 40 epochs with mean geodesic distance $d$ as the loss function and an exclusive subset of the training dataset as the validation set for early stopping. 

\textbf{Data Augmentation.} With a probability of 0.5, we randomly rotate the image clockwise by a degree between 0 and 90, and with another probability of 0.3, we randomly flip the image along an axis that is between 0 and 90 degrees counter-clockwise from the positive x axis. For pixel-wise augmentation, we deploy the popular techniques including translation, resizing, down-sampling for low resolution, hue change, sharpness, grayness, contrast limited adaptive histogram equalization, and brightness. Furthermore, we conduct the following pixel-wise transformations: RGB shift, channel shuffle, gamma correction, color jitter, Gaussian noise, and Gaussian blur. For pixel removal, we utilize center crop and coarse dropout. These methods alter the pixel values, reinforcing a more robust training outcome. 

\textbf{Baseline Models and Evaluation Metric.} We consider many existing HPE models as our baseline models. FSA-Net \cite{yang2019fsa} utilized feature aggregation and soft stagewise regression. HopeNet \cite{Ruiz_2018_CVPR_Workshops} applied multiple losses on a convolutional neural network. TokenHPE \cite{zhang2023tokenhpe} leveraged a transformer to learn the relationship within the facial part. 6DRepNet \cite{hempel2024toward} relied on the geodesic loss function and its variant, 6DRepNet360 \cite{hempel2024toward}, expanded to full range. WHENet \cite{zhou2020whenet}, the other full range model, wrapped the loss function to stabilize the learning of large yaw angles. Following these works, we evaluate the models with mean absolute error (MAE) on each of yaw, pitch, and roll and compute the average of the three MAEs as the Mean. Since no models, including ours, are specialized in one of yaw, pitch, or roll predictions, we mainly compare the Mean. 

\section{Empirical Results}

\subsection{Main Evaluation}
We present the evaluation of all models on six datasets in Table \ref{datasets}. On AFLW2000, CLERF performed nearly on par with 6DRepNet, down by 1.4 degrees in Mean. On BIWI, CLERF performed on par with TokenHPE, falling short by 0.01 degree in Mean. We recognize that having FR coverage, CLERF optimizes for the entire range, leaving relatively less focus on the frontal range that non-full range models directed all of their resources to learn. We act on this to construct a more comprehensive analysis by slightly rotating and flipping the same test images (SA AFLW2000 and SA BIWI), keeping them within the same frontal range that non-full range models trained on. Results show that CLERF outperformed all baseline models, a sign that existing models might have been fixated on the exact angles of the original test images. With 0.57 and 1.29 degrees better than the runner-up for SA AFLW2000 and SA BIWI, respectively, CLERF demonstrates robustness and superiority. Additionally, we examine the full range capability by rotating and flipping the test images on a larger scale and only compare CLERF to full range baseline models. CLERF remains the highest performing, leading by more than 10 degrees in Mean. An interestingly observation is that CLERF had a minor improvement of 0.03 degrees in Mean on FA AFLW2000 compared to the original version, an evidence that CLERF optimizes for the entire range.

\subsection{Ablation Studies}

\begin{table}
\centering
\caption{Supervised learning against contrastive learning. We validate the contrastive learning strategy by comparing the result to that of supervised learning and observe a significant improvement. The optimal results are highlighted in bold.}
\begin{tabular}{c|c|c|c|c|c|c|c|c}
 \toprule
\multirow{2}{*}{Model}  & \multicolumn{4}{|c|}{AFLW} & \multicolumn{4}{|c}{BIWI}\\ 
\cmidrule(lr){2-5} \cmidrule(lr){6-9}
   & Yaw & Pitch & Roll & Mean & Yaw & Pitch & Roll & Mean \\
   \midrule
 CLERF-Supervised & 4.65 & 6.48 & 4.81 & 5.31 & 5.56 & 6.21 & \textbf{2.58} & 4.78 \\
 CLERF  & \textbf{4.22} & \textbf{6.18} & \textbf{4.67} & \textbf{5.02} & \textbf{3.57} & \textbf{4.49} & 3.13 & \textbf{3.73} \\ 
 \bottomrule
\end{tabular}
\label{supervised}
\end{table}

\begin{table}
\centering
\caption{Ablation study on flip and rotation as data augmentation methods. Each of flip and rotation significantly improves the test performance on both the original and SA test images. The integration of flip and rotation achieved the best outcome. The optimal results are highlighted in bold. (Imp. \% indicates the percentage of improvement in Mean.)}
\begin{tabular}{l|c|c|c|c|c|c|c|c}
 \toprule
   Augmentation & \multicolumn{2}{|c|}{AFLW2000} & \multicolumn{2}{|c|}{SA AFLW2000} & \multicolumn{2}{|c|}{BIWI} & \multicolumn{2}{|c}{SA BIWI}\\ 
   \cmidrule(lr){2-3}\cmidrule(lr){4-5}\cmidrule(lr){6-7}\cmidrule(lr){8-9}
   Method & Mean & Imp. \% & Mean & Imp. \% & Mean & Imp. \% & Mean & Imp. \%\\
   \midrule
 CLERF & 6.88 & - & 9.93 & - & 5.48 & - & 8.99 & -\\
 + Flip only & 5.98 & 13\% & 7.78 & 22\% & 4.96 & 9\% & 7.08 & 21\%\\
 + Rotate only & 5.66 & 18\% & 6.19 & 38\% & 4.64 & 15\% & 5.77 & 36\%\\
 + Rotate \& Flip & \textbf{5.02} & \textbf{27\%} & \textbf{5.36} & \textbf{46\%} & \textbf{3.73} & \textbf{32\%} & \textbf{5.10} & \textbf{43\%}\\
  
  \bottomrule
\end{tabular}
\label{ablation_aug}
\end{table}

\begin{figure}[t]
\small
\centering
\begin{minipage}{0.3\linewidth}
\centering
\includegraphics[width=\linewidth]{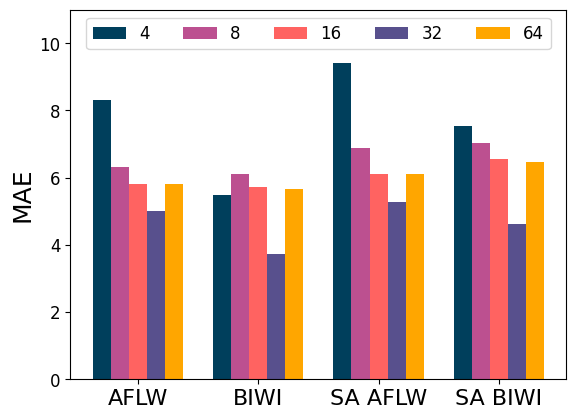}
\centering\mbox{\small(a) Batch size}
\end{minipage}
\begin{minipage}{0.3\linewidth}
\centering
\includegraphics[width=\linewidth]{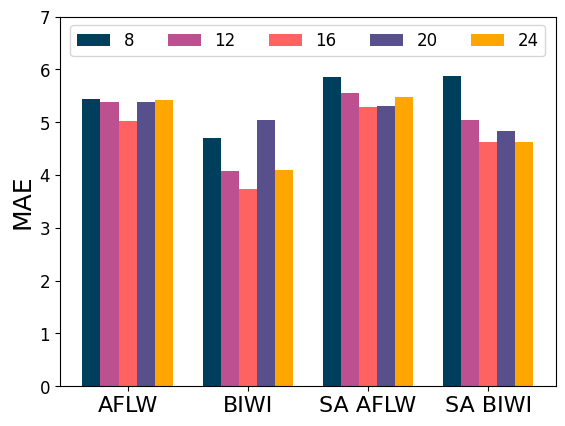}
\centering\mbox{\small(b) Epochs}
\end{minipage}
\begin{minipage}{0.3\linewidth}
\centering
\includegraphics[width=\linewidth]{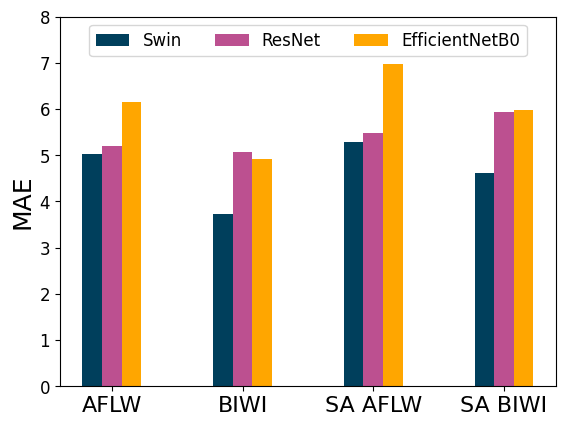}
\centering\mbox{\small(c) Backbone model}
\end{minipage}
\caption{Change of model performance across different choices in (a) batch size, (b) epochs, and (c) backbone model. A batch size of 32, 16 epochs, and Swin Transformer make the best combination. (AFLW2000 is abbreviated as AFLW.)}
\label{fig:ablations}
\end{figure}

\textbf{Contrastive Learning.} We validate the contrastive learning motivation by comparing against the traditional supervised learning approach. We concatenate the representation model with the downstream MLP and train on geodesic loss with the same dataset. The comparison result is shown in Table \ref{supervised}, and we observe a significant improvement with our contrastive learning approach. This demonstrates that contrasting between neighboring angles and further angles boosts the learning of head orientations. By first grouping up the neighboring head poses and separating those that are further away, we enable the downstream MLP to more accurately project the representation to fine-grained head poses.

\textbf{Batch Size.} Batch size is a crucial hyperparameter in contrastive learning when anchor-positives and anchor-negatives are identified within a mini-batch. With sparsity in head orientations, a large batch size tends to increase the number of anchor-positives and the ratio of anchor-positives to anchor-negatives. Figure \ref{fig:ablations}(a) shows the model performance across different batch sizes. We observe a consistent pattern across all four datasets: a larger batch size results in better performance up until a batch size of 32, which corresponds to 32 images from 300W-LP and 32 PanoHead images generated on-the-fly.

\textbf{Epochs.} As the representation model aims to learn a good representation, we study how the number of training epochs affects the learned representation, with results shown in Figure \ref{fig:ablations}(b). Model performance peaked at 16 epochs and declined afterwards. This is likely due to the limited varieties appearing in the synthetic images generated through PanoHead that accumulated over the number of epochs.

\textbf{Image Encoder Backbone.} While we select the Swin Transformer as our image encoder for many of its strengths, we compare to two other backbone models to analyze our proposed approach across different image encoders, as shown in Figure \ref{fig:ablations}(c). The Swin Tranformer consistently outperformed others, with ResNet50 \cite{he2016deep} generally performing better than the EfficientNetB0 model \cite{tan2019efficientnet}, which aligns with their descending number of model parameters.

\textbf{Rotation and Flip Augmentation.} An important component in our approach that enables FRHPE is the rotation and flip as data augmentation techniques. To quantify their significance, we present the respective improvements in Table \ref{ablation_aug}. The combination of rotation and flip performed the best, with rotation or flip alone making their respective contributions. These geometric transformations resulting in better performance not only in AFLW2000 and BIWI, but also in their SA versions.

\subsection{Case Studies and Visualization}
\begin{figure}[t]
\begin{minipage}[t]{0.18\textwidth}
    \centering
    \includegraphics[width=0.9\textwidth]{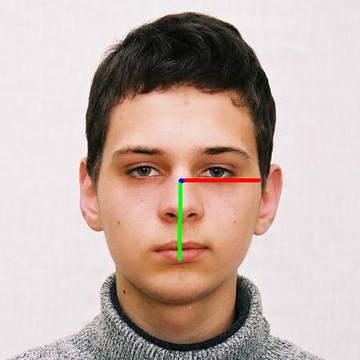}
\end{minipage}
\hfill
\begin{minipage}[t]{0.18\textwidth}
    \centering
    \includegraphics[width=0.9\textwidth]{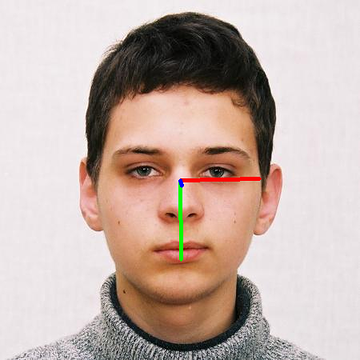}
\end{minipage}
\hfill
\begin{minipage}[t]{0.18\textwidth}
    \centering
    \includegraphics[width=0.9\textwidth]{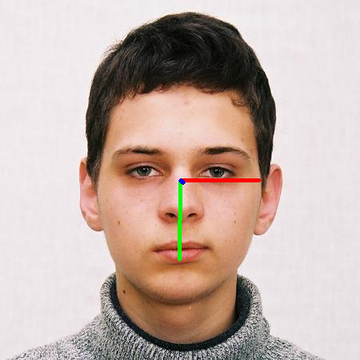}
\end{minipage}
\hfill
\begin{minipage}[t]{0.18\textwidth}
    \centering
    \includegraphics[width=0.9\textwidth]{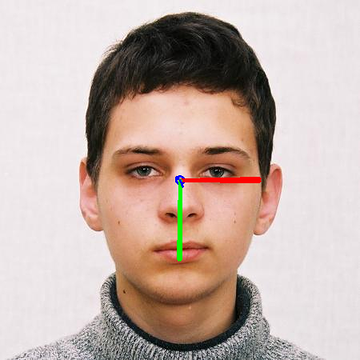}
\end{minipage}
\hfill
\begin{minipage}[t]{0.18\textwidth}
    \centering
    \includegraphics[width=0.9\textwidth]{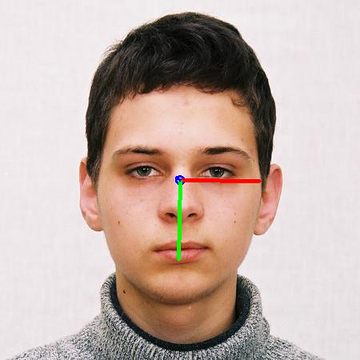}
\end{minipage}

\begin{minipage}[t]{0.18\textwidth}
    \centering
    \includegraphics[width=0.9\textwidth]{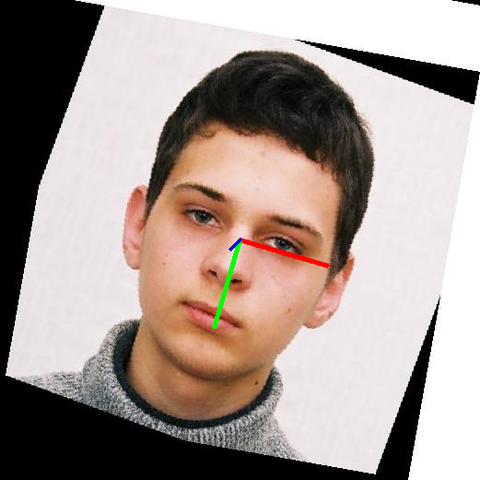}
\end{minipage}
\hfill
\begin{minipage}[t]{0.18\textwidth}
    \centering
    \includegraphics[width=0.9\textwidth]{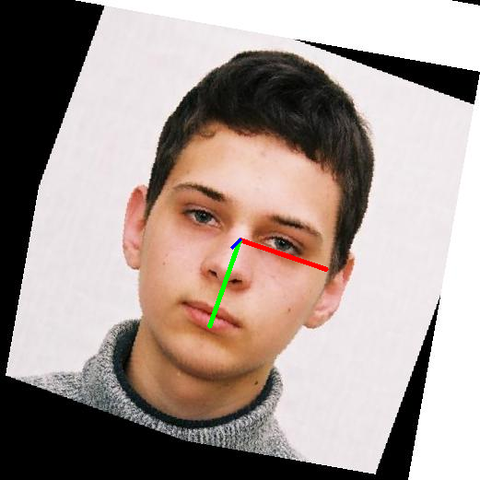}
\end{minipage}
\hfill
\begin{minipage}[t]{0.18\textwidth}
    \centering
    \includegraphics[width=0.9\textwidth]{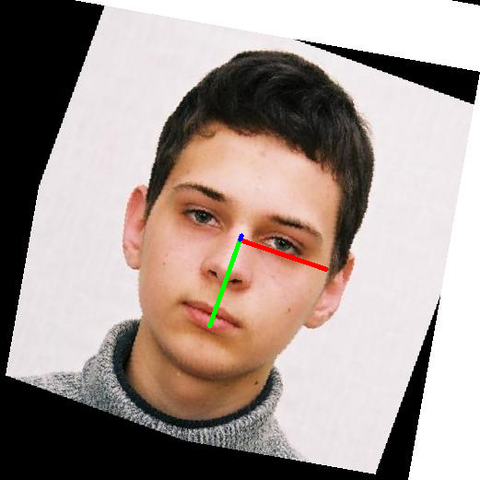}
\end{minipage}
\hfill
\begin{minipage}[t]{0.18\textwidth}
    \centering
    \includegraphics[width=0.9\textwidth]{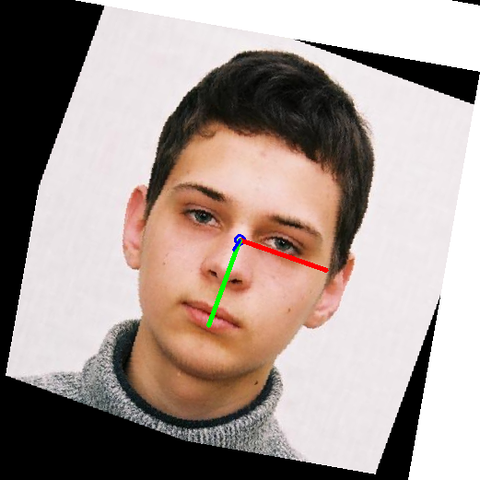}
\end{minipage}
\hfill
\begin{minipage}[t]{0.18\textwidth}
    \centering
    \includegraphics[width=0.9\textwidth]{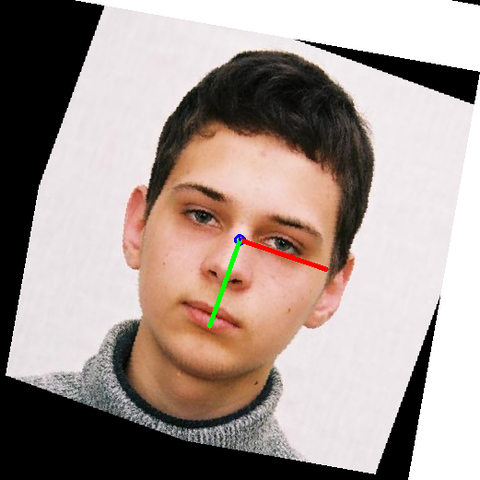}
\end{minipage}

\begin{minipage}[t]{0.18\textwidth}
    \centering
    \includegraphics[width=0.9\textwidth]{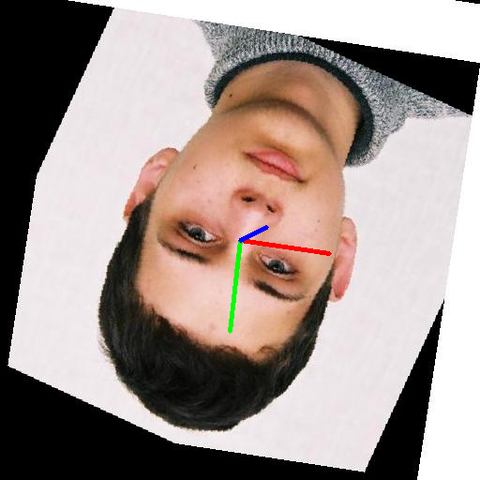}
    \mbox{(a) WHENet}
\end{minipage}
\hfill
\begin{minipage}[t]{0.18\textwidth}
    \centering
    \includegraphics[width=0.9\textwidth]{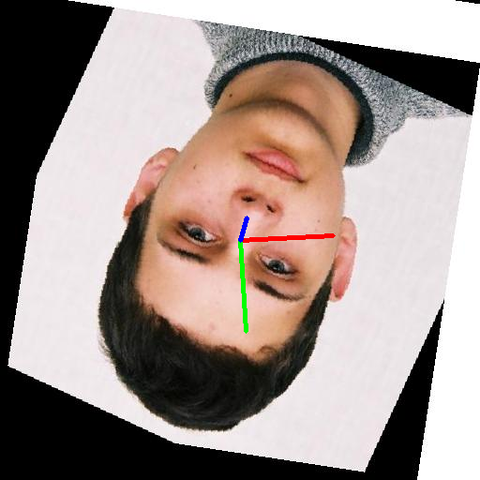}
    \mbox{(b) TokenHPE}
\end{minipage}
\hfill
\begin{minipage}[t]{0.18\textwidth}
    \centering
    \includegraphics[width=0.9\textwidth]{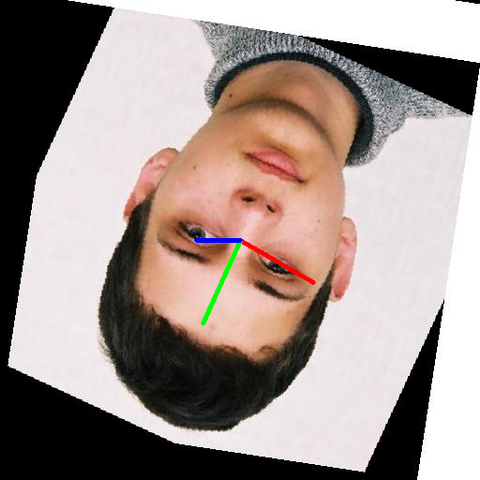}
    \mbox{(c) 6dRepNet360}
\end{minipage}
\hfill
\begin{minipage}[t]{0.18\textwidth}
    \centering
    \includegraphics[width=0.9\textwidth]{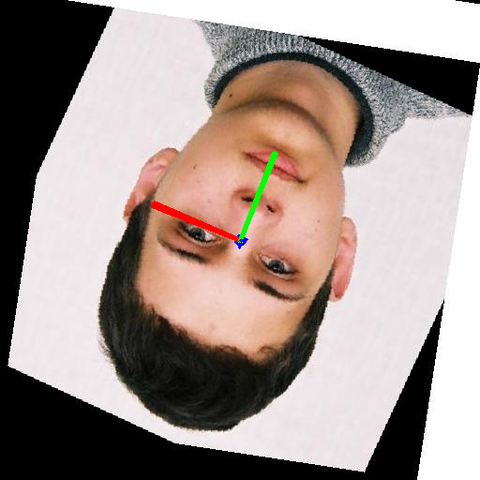}
    \mbox{(d) CLERF (Ours)}
\end{minipage}
\hfill
\begin{minipage}[t]{0.18\textwidth}
    \centering
    \includegraphics[width=0.9\textwidth]{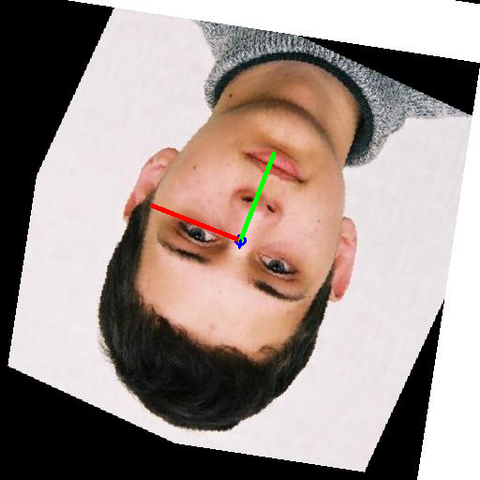}
    \centering\mbox{(e) Ground Truth}
\end{minipage}
\caption{Head pose test predictions of (d) CLERF and (a)$\sim$(c) three baseline models versus the (e) ground truth on the original image (first row) and its SA (second row) and FA (third row) versions. Head pose is represented by the three lines colored in red, blue, and green. We observe barely noticeable differences between the model predictions in the original and SA test image, but CLERF much more accurately predicted the FA head pose.}
\label{fig:cases}
\end{figure}

\textbf{Qualitative Analysis.} After the empirical results, we visually illustrate our model performance with three test cases, each representing one from the original, SA, and FA versions of AFLW2000 test set. We compare to the ground truth and baseline models, as shown in Figure \ref{fig:cases}. The first row presents an original image with very similar head pose predictions across all models, which is consistent with the similar Mean values listed in Table \ref{datasets}. Second row presents the SA version of the same image and we observe more noticeable deviation for WHENet and TokenHPE on the blue line. As for the FA version of the original image shown in the third row, CLERF fully adapted to the heavy rotation while the baseline models struggled, despite that 6dRepNet360 and WHENet aimed at being full range models. This illustrates that although CLERF is on average one degree behind the baseline models on the original AFLW2000 test set, CLERF adapts and produces accurate results in the full range, making the one degree negligible, not to mention the other test set, BIWI, where CLERF performed on par with baseline models on the original images and outperformed them in the SA and FA versions.

\begin{figure}[!t]
\small
\centering
\begin{minipage}{0.18\linewidth}
\centering
\includegraphics[width=0.95\linewidth]{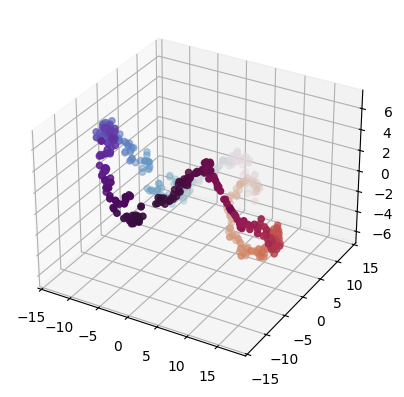}
\centering\mbox{\small(a) WHENet}
\end{minipage}
\begin{minipage}{0.18\linewidth}
\centering
\includegraphics[width=0.95\linewidth]{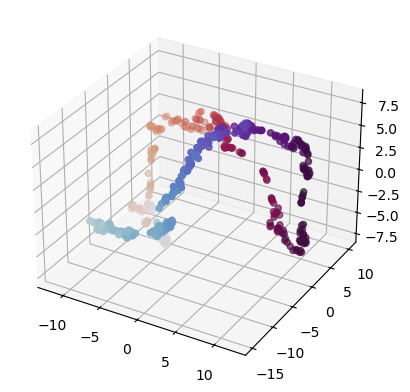}
\centering\mbox{\small(b) 6dRepNet}
\end{minipage}
\begin{minipage}{0.18\linewidth}
\centering
\includegraphics[width=0.95\linewidth]{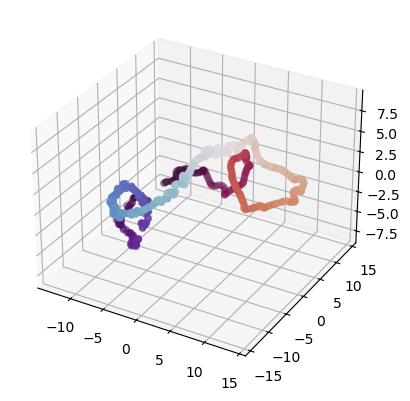}
\centering\mbox{\small(c) 6dRepNet360}
\end{minipage}
\begin{minipage}{0.18\linewidth}
\centering
\includegraphics[width=0.95\linewidth]{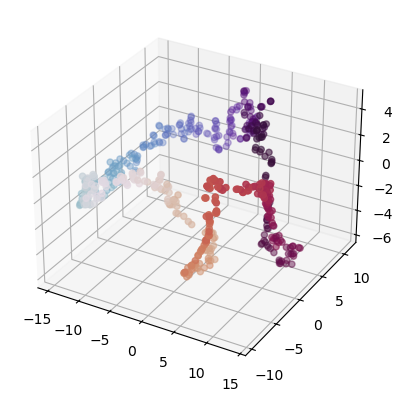}
\centering\mbox{\small(d) CLERF-Supervised}
\end{minipage}
\begin{minipage}{0.17\linewidth}
\centering
\includegraphics[width=1.1\linewidth]{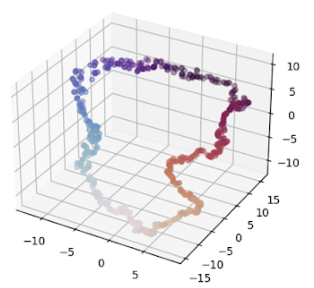}
\centering\mbox{\small(e) CLERF}
\end{minipage}
\caption{3D TSNE plot of the embedding vectors of a video across (a)$\sim$(e) different models. The video shows a person turning around for a total of 360 degrees. The similarity of colors indicates their temporal proximity.}
\label{fig:tsne}
\end{figure}

\textbf{TSNE Visualization.} For comprehensive analysis, we take a further step to visualize with 3D t-distributed stochastic neighbor embedding (TSNE) the embedding vectors of a sequence of images showing a person turning in a full circle. Figure \ref{fig:tsne} compared to the supervised and other baseline models, using a cyclic color map that visualizes neighboring frames with similar colors. An optimal pattern consists of a clear circle with the colors following the cyclic order. CLERF (Figure \ref{fig:tsne}(e)) exhibited clear separation for angles that are further away and kept nearby angles close as shown by the continuous change of colors. On the other hand, lacking only the contrastive learning component, the supervised model (Figure \ref{fig:tsne}(d)) showed the continuity in color changes but did not separate opposite angles as effectively. We made a similar observation with the other baseline models (in Figure \ref{fig:tsne}(a)$\sim$(c)).

\section{Conclusion}
In head pose estimation, the sparsity of head poses has hindered the use of contrastive learning. We proposed a novel idea to tackle this issue and utilize contrastive learning to separate the representation of opposite angles, thereby improving head pose estimation. By generating anchor-positives through a 3D-aware generative adversarial network and the proper geometric transformations, we achieved on par performance with current state-of-the-art models. Though many previous works optimized their performance on the frontal range, we observed a significant decline in performance when the test images are slightly rotated or flipped. In this realm, our model outperformed all baseline models. Furthermore, we validated our full range capability on heavily-transformed images and observed superior performance over other full range models.

\bibliographystyle{plain}
\bibliography{main.bib}


\appendix

\section{Proof of Theorem \ref{theorem1} }
\label{appendix:appendix_theoren_keep_geodesic_dist}
\begin{proof}
Suppose A and B are rotation matrices $\in SO(3)$ and $\theta \leq 90\degree$. According to Equation \ref{formula:flip}, flipping across $L_{\theta}$ on $A$ and $B$ are $A_{flip}(\theta) = Flip{\_}\theta \times A \times Flip{\_}X$ and $B_{flip}(\theta) = Flip{\_}\theta \times B \times Flip{\_}X$. From Equation \ref{gedesic_dist}, $tr(A \times B^{T})$ is the unique factor that can alter the geodesic distance.  To show that the geodesic distance between $A_{flip}(\theta)$ and $B_{flip}(\theta))$ equals to the geodesic distance between $A$ and $B$, it suffices to show the equality $tr(A \times B^{T}) = tr(A_{flip}(\theta) \times B_{flip}(\theta)^{T})$ holds. Let's consider $A_{flip}(\theta) \times B_{flip}(\theta)^{T}$ first.
    \begin{equation}
      \label{proof_equal_trace1}
      \begin{split}
        &A_{flip}(\theta) \times B_{flip}(\theta)^{T} \\
        &= (Flip{\_}\theta \times A \times Flip{\_}X)  \times (Flip{\_}\theta \times B \times Flip{\_}X)^{T} \\
        &= Flip{\_}\theta \times A \times Flip{\_}X  \times Flip{\_}X^{T} \times B^{T} \times (Flip{\_}\theta)^{T}     \\
        &= Flip{\_}\theta \times A \times (Flip{\_}X  \times Flip{\_}X^{T}) \times B^{T} \times (Flip{\_}\theta)^{T}  \\
        &= Flip{\_}\theta \times A \times B^{T} \times (Flip{\_}\theta)^{T}
    \end{split}
    \end{equation}
    Next, the commutativity of the trace of a matrix guarantees the following:
    \begin{equation}
      \label{proof_equal_trace2}
      \begin{split}
        &tr(A_{flip}(\theta) \times B_{flip}(\theta)^{T}) 
        = tr(Flip{\_}\theta \times A \times B^{T} \times (Flip{\_}\theta)^{T}) \\
        &= tr(A \times B^{T} \times (Flip{\_}\theta \times Flip{\_}\theta^{T})) \\
        &= tr(A \times B^{T}).
    \end{split}
    \end{equation}
    Therefore, $d(A, B) = d(A_{flip}(\theta), B_{flip}(\theta))$. Hence, flipping preserves the geodesic distance. Similarly, we can apply Equation \ref{formula:rotation} to prove the equality $tr(A, B) = tr(A_{rotated}(\phi), B_{rotated}(\phi))$ holds for the rotations associated with rotating an image by an angle $\phi$ based on $A$ and $B$, so rotation preserves the geodesic distance as well.
\end{proof}

\section{Examples for Anchor-Positive Synthetic Image Generation}
\label{sec:panohead}
\begin{figure}[H]
\begin{minipage}[t]{0.15\textwidth}
    \centering
    \includegraphics[width=\textwidth]{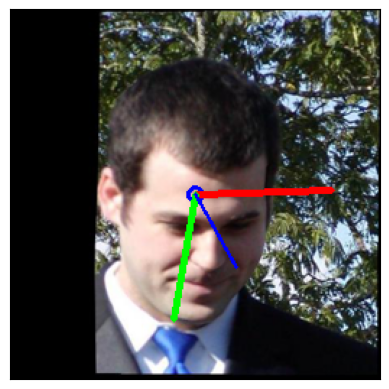}
\end{minipage}
\hfill
\begin{minipage}[t]{0.15\textwidth}
    \centering
    \includegraphics[width=\textwidth]{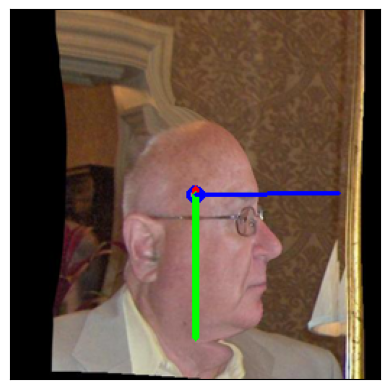}
\end{minipage}
\hfill
\begin{minipage}[t]{0.15\textwidth}
    \centering
    \includegraphics[width=\textwidth]{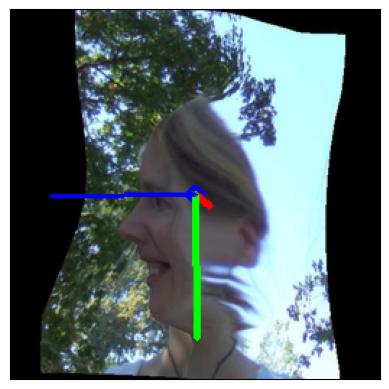}
\end{minipage}
\hfill
\begin{minipage}[t]{0.15\textwidth}
    \centering
    \includegraphics[width=\textwidth]{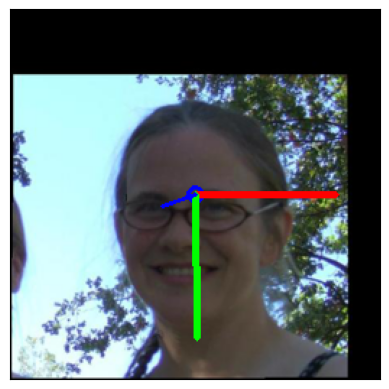}
\end{minipage}
\hfill
\begin{minipage}[t]{0.15\textwidth}
    \centering
    \includegraphics[width=\textwidth]{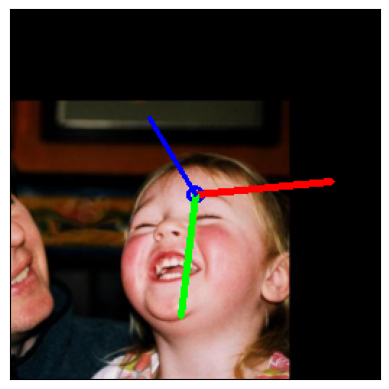}
\end{minipage}
\hfill
\begin{minipage}[t]{0.15\textwidth}
    \centering
    \includegraphics[width=\textwidth]{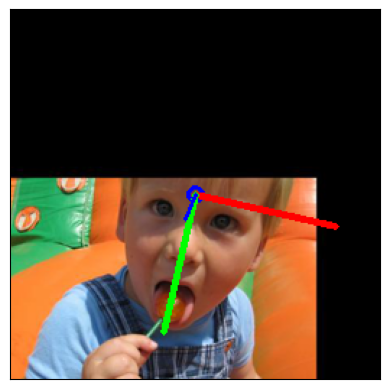}
\end{minipage}

\begin{minipage}[t]{0.15\textwidth}
    \centering
    \includegraphics[width=\textwidth]{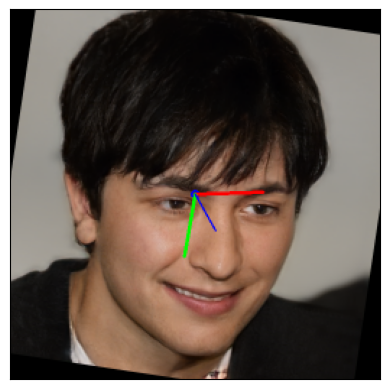}
\end{minipage}
\hfill
\begin{minipage}[t]{0.15\textwidth}
    \centering
    \includegraphics[width=\textwidth]{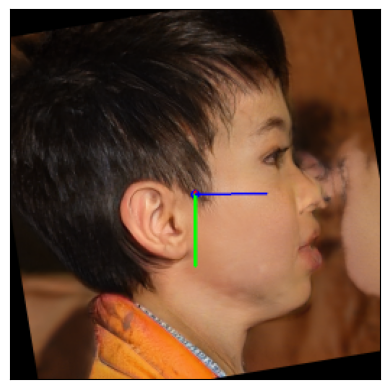}
\end{minipage}
\hfill
\begin{minipage}[t]{0.15\textwidth}
    \centering
    \includegraphics[width=\textwidth]{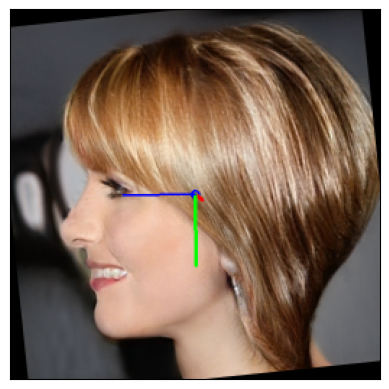}
\end{minipage}
\hfill
\begin{minipage}[t]{0.15\textwidth}
    \centering
    \includegraphics[width=\textwidth]{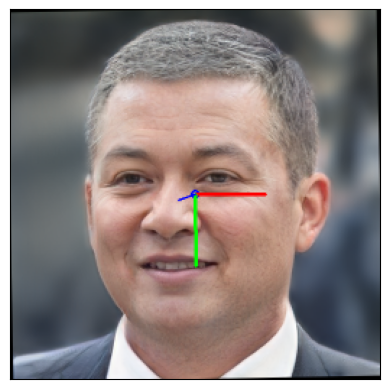}
\end{minipage}
\hfill
\begin{minipage}[t]{0.15\textwidth}
    \centering
    \includegraphics[width=\textwidth]{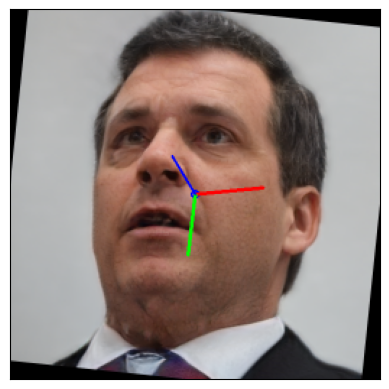}
\end{minipage}
\hfill
\begin{minipage}[t]{0.15\textwidth}
    \centering
    \includegraphics[width=\textwidth]{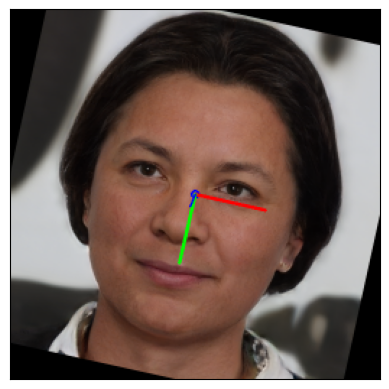}
\end{minipage}

\caption{The first row shows the rotation matrices of 300W-LP dataset images. The second row shows that we can create synthetic images of the same head poses as the first row. In a contrastive learning scenario, the first row are the anchors and the second row are the anchor-positives while anchor-negatives can be selected easily due to the sparsity of head poses.}
\label{fig:panohead}
\end{figure}

\end{document}